\documentclass[runningheads]{llncs}
\usepackage{graphicx}
\usepackage[shortcuts,acronym]{glossaries}
\usepackage[inline]{enumitem}
\usepackage{booktabs}
\usepackage{subcaption}
\usepackage{amsmath}
\usepackage{amssymb}
\usepackage{hyperref}
\usepackage{placeins}

\DeclareMathOperator*{\argmin}{arg\,min \;}
\DeclareMathOperator*{\argmax}{arg\,max \;}

\newcommand*{\name}[1]{\textsc{#1}}

\newcommand{\ul}[1]{\underline{#1}}

\newcommand{\ie}{i.e.\,}
\newcommand{\eg}{e.g.\,}

\newacronym{AutoML}{AutoML}{Automated machine learning}
\newacronym{CASH}{CASH}{combined algorithm selection and hyperparameter optimization}
\newacronym{DAG}{DAG}{directed acyclic graph}
\newacronym{HTN}{HTN}{hierarchical task network}
\newacronym{HPO}{HPO}{hyperparameter optimization}
\newacronym{MCTS}{MCTS}{Monte Carlo tree search}
\newacronym{ML}{ML}{machine learning}
\newacronym{PSO}{PSO}{pipeline synthesis and optimization}

\begin{document}
\title{
	Incremental Search Space Construction for Machine Learning Pipeline Synthesis
}
\titlerunning{Incremental Search Space Construction for ML Pipeline Synthesis}

\author{Marc-Andr\'e Z\"oller\inst{1} \and
Tien-Dung Nguyen\inst{2} \and
Marco F. Huber\inst{3,4}}

\institute{USU Software AG, R\"uppurrer Str. 1, Karlsruhe, Germany
\email{marc.zoeller@usu.com} \and
University of Technology Sydney, Sydney, Australia
\email{TienDung.Nguyen-2@student.uts.edu.au} \and
Institute of Industrial Manufacturing and Management IFF, University of Stuttgart \and
Center for Cyber Cognitive Intelligence CCI, Fraunhofer IPA, Nobelstr. 12, Stuttgart, Germany
\email{marco.huber@ieee.org}}

\authorrunning{M. Z\"oller et al.}

\maketitle

\begin{abstract}
\ac{AutoML} aims for constructing \ac{ML} pipelines automatically. Many studies have investigated efficient methods for algorithm selection and hyperparameter optimization. However, methods for \ac{ML} pipeline synthesis and optimization considering the impact of complex pipeline structures containing multiple preprocessing and classification algorithms have not been studied thoroughly. In this paper, we propose a data-centric approach based on meta-features for pipeline construction and hyperparameter optimization inspired by human behavior. By expanding the pipeline search space incrementally in combination with meta-features of intermediate data sets, we are able to prune the pipeline structure search space efficiently. Consequently, flexible and data set specific \ac{ML} pipelines can be constructed. We prove the effectiveness and competitiveness of our approach on \(28\) data sets used in well-established \ac{AutoML} benchmarks in comparison with state-of-the-art \ac{AutoML} frameworks.

\end{abstract}

\section{Introduction}

\ac{AutoML} promises to automate the synthesis of \ac{ML} pipelines, handling \ac{HPO}, algorithm selection and pipeline structure search. Many publications have proven the superiority of Bayesian optimization for \ac{HPO} and algorithm selection, formulated as \ac{CASH}, over classic approaches like grid or random search \cite{Bergstra2013,Hutter2011}. More recently, methods for composing complete \ac{ML} pipelines from a set of algorithms have been proposed. Those methods have a holistic view on pipeline synthesis: pipeline structure search is considered as an extension of \ac{CASH} where, instead of a single algorithm, a combination of multiple algorithms is selected and optimized simultaneously. Due to the exponential growth in complexity, the pipeline structure space is usually not evaluated thoroughly.

In contrast to current \ac{AutoML} approaches, data scientists often create an \ac{ML} pipeline in several small distinct steps. Starting from an empty pipeline, data scientists add algorithms to an \ac{ML} pipeline incrementally by taking a detailed look at how the data set evolves in a pipeline in combination with their profound experience. Only if a pipeline structure performs well enough, a fine-tuning via hyperparameters is performed.

In this paper, we propose an alternative data-centric view on pipeline structure synthesis inspired by human behavior that allows an adaption of a pipeline to a specific data set. Through extensive use of meta-learning, we are able to dynamically prune the pipeline structure search space depending on meta-features of intermediate data sets. Intermediate data sets are the outputs of the execution of each individual step in a pipeline. Furthermore, the \ac{HPO} of a pipeline candidate is boosted via knowledge-sharing between different pipelines. The main contributions of this paper are as follows:
\begin{itemize}
    \item We reformulate the \ac{CASH} and pipeline synthesis problem to enable efficient measures to reduce the pipeline search space and warm-starting \ac{CASH}.
	\item We present a data-centric approach for incremental pipeline synthesis and hyperparameter optimization without expert knowledge inspired by human behavior called \name{dswizard}.
	\item To ensure reproducibility of the results, we publish our meta-learning base consisting of \(13.5\) million unique \ac{ML} pipelines on \(28\) data sets.
\end{itemize}

In Section~\ref{sec:related_work} related work regarding pipeline structure synthesis and meta-learning in \ac{AutoML} is discussed. Section~\ref{sec:methods} describes how we model the pipeline synthesis and the creation of the meta-learning base. The effectiveness of this approach is evaluated in Section~\ref{sec:experiments} followed by a short conclusion in Section~\ref{sec:conclusion}.

\section{Preliminary and Related Work}
\label{sec:related_work}

Let a classification task---containing a data set \(D = \{ (\vec{x}_1,y_1), \dots,(\vec{x}_m,y_m) \} \) with \(\vec{x}_i \in \mathbb{X}^{d}\) being the input domain and \(y \in \mathbb{Y} \subset N\) the target domain and a loss function \(\mathcal{L}: \mathbb{Y}^2 \rightarrow \mathbb{R} \)---be given. Furthermore, let a fixed set of algorithms be given as \(\mathcal{A} = \left\{ A^{(1)}, A^{(2)}, \dots, A^{(n)} \right\} \). Each algorithm \(A^{(i)}\) is a transformation \(\phi: \mathbb{Z} \rightarrow \mathbb{Z}'\) between two arbitrary domains. In case of \(\mathbb{Z}' = \mathbb{Y}\) we denote the algorithm as a \textit{classifier}, otherwise as a \textit{preprocessor}. Usually \(A^{(i)}\) is configured by hyperparameters \(\vec{\lambda}^{(i)}\) from a domain \(\Lambda_{A^{(i)}}\). \(A^{(i)}\) transforming a data set \(D\) being configured by \(\vec{\lambda}\) is denoted as \(\phi^{(i)}_{\vec{\lambda}}(D)\). An \ac{ML} pipeline \(\mathcal{P}\) is a sequential combination of algorithms mapping data from an input domain to a target domain \(f_{\mathcal{P}}: \mathbb{X}^d \rightarrow \mathbb{Y}\). It consists of a pipeline structure \(g\)---usually modeled as a \ac{DAG}---with length \(|g|\), algorithms \(\vec{A} = [A_1, \dots, A_{|g|}]\) and the according hyperparameters \(\vec{\lambda} = [\Lambda_1, \dots, \Lambda_{|g|}]\). \ac{AutoML} aims at generating a pipeline \(\mathcal{P}\) that optimizes
\begin{equation*}
	(g, \vec{A}, \vec{\lambda})^* \in \argmin_{g \in G, \vec{A} \in \mathcal{A}^{|g|}, \vec{\lambda} \in \Lambda_{A_1} \times \dots \times \Lambda_{A_{|g|}} } \pi \left(g, \vec{A}, \vec{\lambda}, D \right)
\end{equation*}
with
\begin{equation*}
	\pi \left(g, \vec{A}, \vec{\lambda}, D \right) = \dfrac{1}{m} \sum_{i = 1}^m \mathcal{L}(\hat{y}_{i}, y_{i})
\end{equation*}
with \(\hat{y}_{i}\) being the predicted output on the sample \(\vec{x}_i\). We refer to this extension of the \ac{CASH} problem as \ac{PSO} problem.

The \ac{CASH} notation, as originally introduced by \cite{Thornton2013}, extends \ac{HPO} by introducing an additional categorical meta-hyperparameter that represents an algorithm choice. This approach does not scale well as the search space grows exponentially with the length of the pipeline \cite{Kandasamy2015}. To counter this problem, many frameworks use a fixed pipeline structure based on best-practices reducing \ac{PSO} to \ac{CASH}, \eg \cite{Komer2014,Swearingen2017}. \name{autosklearn} \cite{Feurer2015} allows the omission of single steps in a fixed pipeline, effectively replacing a fixed pipeline structure with a small set of pipeline candidates. Similarly, \name{P4ML} \cite{Gil2018} uses a set of hand-crafted, best-practice pipelines for a wide variety of task instances. Appropriate pipeline candidates are selected based on data set meta-features followed by a fine-tuning via \ac{HPO}. Yet, even when selecting from a set of fixed structures, the pipeline structure cannot be freely adapted to a specific problem instance.

\name{TPOT} \cite{Olson2016} uses genetic programming to solve the \ac{PSO} problem. \name{RECEIPE} \cite{DeSa2017} extends \name{TPOT} by incorporating a context-free grammar to guide the construction of pipeline structures. Even though this approach is able to build flexible tree-shaped pipelines, experiments have shown that genetic programming approaches tend to build pipelines using only one or two algorithms \cite{Zoller2019}.

Multiple approaches that use a \ac{MCTS} \cite{Coulom2006} for pipeline synthesis have been proposed. \name{ML-Plan} \cite{Mohr2018} traverses a hierarchical task network with a \ac{MCTS} to perform \ac{PSO}. By design, the structure is determined first followed by the \ac{HPO}. To assess the score of incomplete pipelines, random path completion is used, which does not scale well to high dimensions \cite{Hutter2018}. Similarly, \name{AlphaD3M} \cite{Drori2019} uses a combination of \ac{MCTS} and neural networks to build pipeline structures based on a grammar while ignoring \ac{HPO} completely. These approaches are more flexible in comparison to semi-fixed pipelines but still enforce specific pipeline patterns.

Many \ac{AutoML} approaches use meta-learning to warm-start \ac{CASH} or find promising pairs of preprocessors and classifiers \cite{Hutter2018}. \name{AlphaD3M} uses meta-features and the algorithms in the current pipeline to predict the performance of a possible pipeline candidate. However, all those approaches only calculate meta-features of the initial data set.

\section{\name{DSWIZARD} Methodology}
\label{sec:methods}

\begin{figure}
\centering
\includegraphics[width=\textwidth]{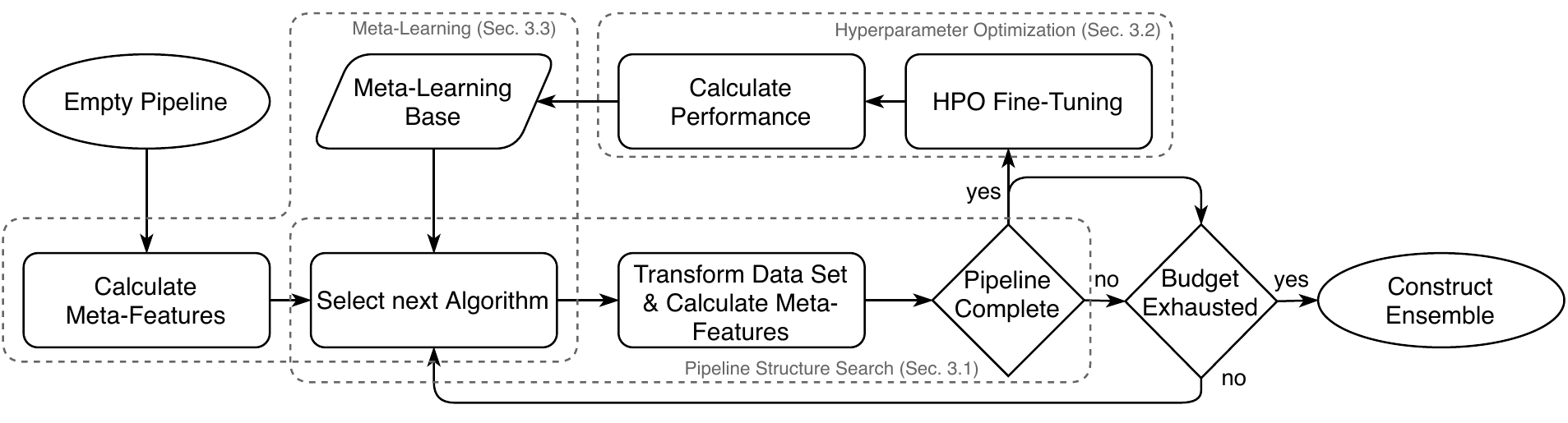}

\caption{General pipeline synthesis procedure in \name{dswizard}.}
\label{fig:procedure}
\end{figure}

Our approach, dubbed \name{dswizard}, is inspired by the behavior of a human data scientist creating an \ac{ML} pipeline. Starting from a very basic pipeline with default hyperparameters, the pipeline structure is extended gradually based on the characteristics of the intermediate data sets and the experience from previous tasks. After obtaining a combination of algorithms, a fine-tuning of the hyperparameters is performed. Fig.~\ref{fig:procedure} contains an overview of our proposed approach.

Starting from an empty pipeline and a given data set, at first the meta-features of the data set are extracted. Based on these meta-features, a suited algorithm is selected using information from a pre-computed meta-learning base. Next, the data set is transformed using the selected algorithm and its default hyperparameters. Whenever the pipeline ends in a classifier, a fine tuning via \ac{HPO} is performed to obtain a performance measure. This procedure is repeated until a predefined time budget is exhausted. Finally, ensemble selection \cite{Caruana2004} is used to create an ensemble of the best performing pipelines.

More formally, we reformulate the \acl{PSO} problem as a bilevel optimization problem:

\begin{equation*}
\begin{split}
	(g, \vec{A}, \vec{\lambda}^*)^* \in \argmin_{g \in G, \vec{A} \in \mathcal{A}^{|g|}}	\quad & \pi \left(g, \vec{A}, \vec{\lambda}^*, D \right) \\
	\mathrm{s.t.} 				\quad & \vec{\lambda}^* \in \argmin_{ \vec{\lambda} \in \Lambda_{A_1} \times \dots \times \Lambda_{A_{|g|}} } \pi \left(g, \vec{A}, \vec{\lambda}, D \right).
\end{split}
\end{equation*}

The outer optimization problem is used for pipeline structure search and algorithm selection. The inner optimization problem performs the \ac{HPO} of the selected algorithms. The implementation is publicly available on GitHub.\footnote{
    See \url{https://github.com/Ennosigaeon/dswizard}.
}

\subsection{Incremental Pipeline Structure Search}
As each pipeline has to have a finite length and the set of algorithms \(\mathcal{A}\) is finite, it is possible to enumerate the complete pipeline search space up to a depth \(d\). The resulting search space can be interpreted as a layered \ac{DAG}. Each node/state \(s_t\) represents a pipeline candidate \(g_{s_t}\), \ie, a list of algorithms, and a vector of meta-features of the intermediate data set \(D_{s_t}\) obtained by applying the (incomplete) pipeline candidate to the input data set; each edge/action represents an algorithm \(A_t \in \mathcal{A}\). All nodes in a layer have pipeline candidates with identical lengths. We use an adapted \ac{MCTS} \cite{Coulom2006} to efficiently traverse the search graph. In contrast to all existing solutions, we explicitly calculate the meta-features of each intermediate data set and not only the initial data set.

The policy for \ac{MCTS}'s \textit{selection} and \textit{expansion} phase---inspired by \textit{polynomial upper confidence trees} \cite{Auger2013}---is
\begin{equation*}
	A_t \in \argmax_{A \in \mathcal{A}} o(s_t) \cdot \big( Q(s_t, A) + c(t) \cdot U(s_t, A) \big)
\end{equation*}
weighting the exploitation \(Q\) and exploration \(U\) by a function \(c\) for a given action \(A \in \mathcal{A}\) and state \(s_t\). Exploitation is defined as
\begin{equation*}
	Q(s_t, A) = \dfrac{P(s_t, A)}{1 + N(s_t, A)} \sum_{s' \in s_t(A)} \nu(s')
\end{equation*}
with \(P(s_t, A)\) being a prior performance estimate (see Section~\ref{sec:meta-learnig}), \(N(s_t, A)\) being the number of times action \(A\) was selected in state \(s_t\), \(s_t(A)\) the state obtained after applying action \(A\) and \(\nu(s')\) a previously observed performance in state \(s'\). Exploration, defined as
\begin{equation*}
	U(s_t, A) = \dfrac{\sqrt{\sum_{b \in \mathcal{A}} N(s_t, b)}}{1 + N(s_t, A)}~,
\end{equation*}
calculates the multiplicative inverse of the relative number of selections of \(A\), giving a higher weight to less frequently selected actions. To account for overfitting we introduce an additional exponential term
\begin{equation*}
	o(s_t) = 1 - \dfrac{c^{|g_{s_t}|}}{c^{l_{\mathrm{max}}}}
\end{equation*}
that reduces the node reward depending on the current pipeline length, a maximum pipeline length \(l_{\mathrm{max}}\) and a scaling constant \(c > 1\).

The \ac{MCTS} procedure is adapted such that selection can be aborted if the current node has a higher reward than all child nodes. Similarly, expansion can be skipped. During each expansion step the data set in \(s_t\) is transformed by \(A_t\) and stored. Usually, \ac{MCTS} uses a \textit{simulation} to calculate the reward of \(s_t\). However, a few random simulations do not scale well in high dimensional search spaces and many simulations are prohibitively expensive \cite{Hutter2018}. Instead, expansion is repeated recursively until the pipeline candidate ends in a classifier. After \(e\) consecutive expansions the selection of a classifier is enforced. Conceptually, this is similar to a random simulation. However, as we immediately know the meta-features of each intermediate data set, the simulation can be guided by the meta-learning base explained in Section~\ref{sec:meta-learnig}. This approach explicitly does not restrict the pipeline structure via a grammar or similar measures.

The reward \(\nu(s_{t + 1})\) is not directly obtained during the \ac{MCTS}. Instead it is computed via the \ac{HPO} procedure described in Section~\ref{sec:hpo}. Therefore, pipeline structure search and \ac{HPO} can be solved independently of each other while still being tightly coupled.

Finally, it remains to decide how many \ac{HPO} samples are drawn to determine the reward of a state. To quickly discard unpromising structures, we wrap the complete structure search procedure in a multi-fidelity approach similar to \name{Hyperband} \cite{Li2018}. Yet, instead of deterministically calculated budgets---the number of \ac{HPO} iterations---we adapt the greediness of the policy depending on the remaining optimization time
\begin{equation*}
    c(t) = w \cdot \left( \exp\left(\dfrac{t_\mathrm{max} - t}{t_\mathrm{max}}\right) - 1 \right)~,
\end{equation*}
with \(t_\mathrm{max}\) being the total optimization time and \(w\) a non-negative weighting constant.
For each pipeline candidate a small fixed number of \ac{HPO} iterations is performed leading to more \ac{HPO} iterations on well-performing candidates.

Using this procedure, the pipeline structure search space is incrementally expanded whenever a new layer of the graph is visited. Simultaneously, ineffective---the algorithm does not modify the data set---incompatible or bad performing transformations can be identified quickly. Consequently, the search graph is pruned efficiently, often even without any \ac{HPO}.

\subsection{Hyperparameter Optimization}
\label{sec:hpo}

After fixing the pipeline structure, its actual performance has to be assessed. In general, this step is computationally equally expensive as optimizing the hyperparameter for a fixed pipeline. Consequently, an efficient \ac{HPO} procedure is key to evaluating various pipeline structures.

Traditional \ac{CASH} procedures model the problem for a fixed structure \(g\) with---for simplicity---only one algorithm as 
\begin{equation*}
	\vec{\lambda}^* \in \argmin_{ A^{(i)} \in \mathcal{A}, \vec{\lambda} \in \Lambda_{A^{(i)}} } \pi_g \left(A^{(i)}, \vec{\lambda}, D \right) .
\end{equation*}
selecting the actual algorithm and its hyperparameters simultaneously, configuring all algorithms accordingly and finally evaluating the performance on the input data set. This approach has three major drawbacks:
\begin{enumerate*}
	\item The transformation of the data set being processed is not considered.
	\item The algorithms in a pipeline may be incompatible with each other due to implicit requirements of the used algorithms on the input data. Selecting and fitting all algorithms at once may lead to wasted optimization time as incompatibilities are only detected during fitting \cite{Nguyen2020}.
	\item Sharing knowledge about well performing configurations between different pipeline structures using the same subset of algorithms is not possible.
\end{enumerate*}
Instead we propose, to use a distinct optimization instance 
\begin{equation*}
	\vec{\lambda}_i^* \in \argmin_{\vec{\lambda} \in \Lambda_{A^{(i)}}} \pi \left(A^{(i)}, \vec{\lambda},  D \right)
\end{equation*}
for each algorithm only considering \ac{HPO}. To prevent an improper impact of previous algorithms and their hyperparameter on the optimization, we additionally require that all meta-features of the transient data set \(D\) have to be similar. Otherwise, a new \ac{HPO} instance is created. This allows sharing knowledge about well-performing hyperparameters between identical algorithms in different pipeline candidates, given that the pipeline prefixes yielded similar data sets.

The hyperparameters of each algorithm can be selected ``on-the-fly" in order of appearance of the algorithms. After selecting the hyperparameters for each algorithm, the final performance is back-propagated to update each optimizer leading to a formulation compatible with current \ac{CASH} formulations
\begin{equation*}
	\vec{\lambda}^* \in \argmin_{\vec{\lambda}_1 \in \Lambda_{1}, \dots , \vec{\lambda}_{|g|} \in \Lambda_{|g|} } \pi \left(
	A_{|g|}, \vec{\lambda}_{|g|}, \phi_{\vec{\lambda}_{|g|}}^{(|g|)} \left(
	\phi_{\vec{\lambda}_{|g - 1|}}^{(|g - 1|)} \left(
	\dots
	\phi_{\vec{\lambda}_{1}}^{(1)} (D) \right) \right)\right)
\end{equation*}
with \(\vec{\lambda}^* = \vec{\lambda}_{1} \cup \dots \cup \vec{\lambda}_{|g|}\) and \(\vec{A}\) being provided via the previously described structure search. Consequently, the hyperparameter search space for a single algorithm is significantly smaller than the complete \ac{CASH} search space. This imposes two major benefits:
\begin{enumerate*}
	\item Bayesian optimization methods have been proven to be more powerful for low dimensional search spaces \cite{Kandasamy2015}.
	\item The small search space improves the applicability of warm-starting. Based on the meta-features of the intermediate data set, samples for warm-starting can be extracted from previously evaluated configurations on different pipeline structure candidates.
\end{enumerate*}
Each individual optimization problem can be solved via standard procedures like \name{SMAC} or \name{hyperopt}. In the context of this work tree Parzen estimators \cite{Bergstra2013} are used. Each instance of this procedure yields a new performance measure \(\nu(s_{t + 1})\) for the \ac{MCTS} procedure.

\subsection{Meta-Learning}
\label{sec:meta-learnig}
Traditional \ac{MCTS} uses \textit{simulations} to determine the score of an unvisited node. As extensive simulations are prohibitively expensive in the context of \ac{ML}, current \ac{AutoML} tools use a small number of random completions, potentially restricted by a grammar, to estimate the reward of a state, \eg \cite{Mohr2018}. We propose to guide the random completions by considering intermediate meta-features.

To get a diverse foundation for the meta-learning base, we collected \(30\) unique data sets from \name{OpenML} \cite{Vanschoren2014}. Starting from the input data set, each available algorithm is applied using its default hyperparameters. The transformed data set is added to the meta-learning base---in case of a classifier, the transformed data set consists of the input data set with the prediction added as a new feature. This procedure is repeated exhaustively until the maximum pipeline length of five algorithms is reached. For each data set in the meta learning base, \(40\) meta-features are extracted. As the meta-feature extraction has to be applied in each stage of the \ac{MCTS}, a fast calculation of the meta-features is important, which limits the available meta-features to general, statistical, information-theoretical and simple model-based meta-features.\footnote{The complete list of all data sets, algorithms and used meta-features is available in the online Appendix alongside the source code at \url{https://git.io/JIOaJ}.}

If the applied algorithm comprises a classifier, the current performance is evaluated. For preprocessing algorithms the performance is estimated using all subsequent classification algorithms. Using this approach, we extracted the performance of over \(13.5\) million unique pipelines on \(30\) diverse data sets.

To account for the stochastic environment, the performance prediction of an algorithm for a given state is modeled by a normal distribution
\begin{equation*}
	P(s_t, A) \sim \mathcal{N} \left( RF_\mu(s_t, A), RF_\sigma(s_t, A) \right)
\end{equation*}
with \(RF_\mu\) and \(RF_\sigma\) being two random forest regression models trained on the mean and standard deviation of the performance, respectively. The complete meta-learning base, namely the raw data and trained models, is publicly available alongside the source code but we also plan to publish all pipelines on \name{OpenML}.

\section{Experiments}
\label{sec:experiments}

To prove the effectiveness and efficiency of our approach, \name{dswizard} is compared with the two best established \ac{AutoML} tools: \name{autosklearn} and \name{TPOT}. Additionally, we perform an ablation study in which we test a variation of \name{dswizard} without meta-learning, dubbed \name{dswizard*}, to get a better impression of the impact of meta-learning during structure synthesis on the final performance.

\subsection{Experiment Setup}

To ensure fair and comparable results, the existing \name{OpenML} \ac{AutoML} benchmark framework \cite{Gijsbers2019} is used for all experiments. We reuse the predefined constraints of a \(60\) minute optimization timeout per fold. Experiments are conducted on \textit{e2-standard-4} virtual machines on Google Cloud Platform equipped with Intel Xeon E5 processors with 4 cores and 16 GB memory.

All frameworks are evaluated on \(28\) publicly available binary and multiclass classification data sets from established and curated \ac{AutoML} benchmark suits \cite{Bischl2017,Gijsbers2019}. More specifically, \name{OpenML} tasks are used for each data set. A task provides information about a data set, for example how train-test splits have to be done or which loss function to use, to enable comparable results. The performance of each final configuration is computed using a hold-out validation data set. For binary and multiclass tasks \textit{AUC} and \textit{logloss} are used as metric, respectively.

To eliminate the impact of different search spaces on the final performance, the existing \name{TPOT} and \name{autosklearn} adapters are adopted to use the same search space as \name{dswizard}. This includes the available algorithms, hyperparameters per algorithm as well as search ranges.\footnote{	
    At least if supported by the frameworks. For example, \name{TPOT} can only handle discretized continuous hyperparameters.
} The complete search space consists of \(35\) algorithms with \(38\) categorical and \(62\) numerical hyperparameters. A short summary of the configuration space is provided in the online Appendix.

To prevent a leaking of information via meta-learning in \name{dswizard}, we construct an individual meta-learning base for each data set excluding the data set under evaluation. The \name{OpenML} \ac{AutoML} benchmark's \name{autosklearn} adapter always performs label encoding before passing the data to the optimization routine. Similarly, the \name{TPOT} adapter always performs label encoding and an imputation of missing data. As these algorithms are not obtained via an optimization procedure, they are not further considered.

\subsection{Experiment Results}

Table \ref{tbl:auc_results_cash_solver} and \ref{tbl:logloss_results_cash_solver} contain the final test performances of all evaluations. For each data set, the mean performance and standard deviation over \(10\) folds is reported. Bold face represents the best mean value for each data set. Results not significantly worse than the best result, according to a two-sided Wilcoxon signed-rank test with \(p = 0.05\), are underlined. If a framework consistently failed to yield results for at least three folds, the performance for that data set is not recorded.

\begin{table}
\center

\caption{
	Final test performance for all tested binary classification data sets using \textit{AUC} as metric. Larger values are better.
}
\label{tbl:auc_results_cash_solver}

\newrobustcmd{\B}{\fontseries{b}\selectfont}

\begin{tabular}{@{} l l l l l @{}}
	\toprule
	Data Set		& \name{Autosklearn}		& \name{TPOT}		& \name{dswizard} & \name{dswizard*}\\
	\midrule

Australian       	&        0.9337 \(\pm\) 0.0186	&        0.9380 \(\pm\) 0.0180	& \B     0.9660 \(\pm\) 0.0125	&        0.9461 \(\pm\) 0.0219	\\
ada\_agnostic    	&        0.9046 \(\pm\) 0.0161	&        0.9034 \(\pm\) 0.0136	& \B     0.9138 \(\pm\) 0.0140	&    \ul{0.9121 \(\pm\) 0.0171}	\\
adult            	& \B     0.9279 \(\pm\) 0.0052	&        ---                  	&        0.8933 \(\pm\) 0.0121	&        0.8921 \(\pm\) 0.0103	\\
bank-marketing   	& \B     0.9350 \(\pm\) 0.0068	&        ---                  	&        0.9172 \(\pm\) 0.0302	&        0.9090 \(\pm\) 0.0132	\\
blood-transfusion	&        0.7288 \(\pm\) 0.0544	&    \ul{0.7601 \(\pm\) 0.0664}	& \B     0.7885 \(\pm\) 0.0649	&    \ul{0.7310 \(\pm\) 0.0632}	\\
credit-g         	&        0.7719 \(\pm\) 0.0318	&        0.7731 \(\pm\) 0.0517	& \B     0.8527 \(\pm\) 0.0260	&        0.8050 \(\pm\) 0.0315	\\
eeg-eye-state    	& \B     0.9909 \(\pm\) 0.0033	&        ---                  	&    \ul{0.9903 \(\pm\) 0.0041}	&        ---                  	\\
higgs            	& \B     0.8084 \(\pm\) 0.0061	&        0.7902 \(\pm\) 0.0123	&        0.7263 \(\pm\) 0.0844	&        0.7667 \(\pm\) 0.0345	\\
jasmine          	&        0.8814 \(\pm\) 0.0167	&        0.8880 \(\pm\) 0.0167	& \B     0.9073 \(\pm\) 0.0177	&        0.8977 \(\pm\) 0.0209	\\
kc2              	&        0.8162 \(\pm\) 0.0931	&    \ul{0.8252 \(\pm\) 0.1599}	& \B     0.8911 \(\pm\) 0.0447	&        0.7867 \(\pm\) 0.0522	\\
kr-vs-kp         	&    \ul{0.9992 \(\pm\) 0.0014}	&        0.9975 \(\pm\) 0.0046	& \B     0.9995 \(\pm\) 0.0009	&    \ul{0.9979 \(\pm\) 0.0057}	\\
nomao            	& \B     0.9956 \(\pm\) 0.0009	&        0.9936 \(\pm\) 0.0044	&        0.9939 \(\pm\) 0.0015	&        0.9935 \(\pm\) 0.0018	\\
numerai28.6      	&    \ul{0.5291 \(\pm\) 0.0053}	&    \ul{0.5267 \(\pm\) 0.0042}	& \B     0.5311 \(\pm\) 0.0116	&    \ul{0.5286 \(\pm\) 0.0081}	\\
phoneme          	&    \ul{0.9629 \(\pm\) 0.0101}	&    \ul{0.9656 \(\pm\) 0.0082}	& \B     0.9662 \(\pm\) 0.0086	&    \ul{0.9638 \(\pm\) 0.0087}	\\
sa-heart         	&        0.7586 \(\pm\) 0.0485	&    \ul{0.7551 \(\pm\) 0.1196}	&    \ul{0.7968 \(\pm\) 0.1065}	& \B     0.8321 \(\pm\) 0.0651	\\
sylvine          	&    \ul{0.9899 \(\pm\) 0.0037}	&    \ul{0.9855 \(\pm\) 0.0075}	& \B     0.9903 \(\pm\) 0.0032	&        0.9865 \(\pm\) 0.0042	\\

	\bottomrule
\end{tabular}

\end{table}

\begin{table}[ht!]
\center

\caption{
	Final test performance for all tested multiclass classification data sets using \textit{logloss} as metric. Smaller values are better.
}
\label{tbl:logloss_results_cash_solver}

\newrobustcmd{\B}{\fontseries{b}\selectfont}

\begin{tabular}{@{} l l l l l @{}}
	\toprule
	Data Set		& \name{Autosklearn}		& \name{TPOT}		& \name{dswizard} & \name{dswizard*}  \\
	\midrule

Helena           	& \B     3.0091 \(\pm\) 0.1153	&        ---                  	&    \ul{3.0226 \(\pm\) 0.0829}	&        3.2283 \(\pm\) 0.2200	\\
Jannis           	& \B     0.7016 \(\pm\) 0.0235	&        0.7297 \(\pm\) 0.0417	&        0.7582 \(\pm\) 0.0524	&        0.7861 \(\pm\) 0.0856	\\
Shuttle          	& \B     0.0006 \(\pm\) 0.0004	&        ---                  	&    \ul{0.0011 \(\pm\) 0.0013}	&        ---                  	\\
analcatdata\_auth	&    \ul{0.0691 \(\pm\) 0.1285}	&    \ul{0.0123 \(\pm\) 0.0210}	&    \ul{0.0182 \(\pm\) 0.0378}	& \B     0.0073 \(\pm\) 0.0090	\\
analcatdata\_dmft	&        1.7520 \(\pm\) 0.0233	&        ---                  	& \B     1.7147 \(\pm\) 0.0288	&        1.7498 \(\pm\) 0.0676	\\
car              	&    \ul{0.0045 \(\pm\) 0.0075}	& \B     0.0033 \(\pm\) 0.0035	&    \ul{0.0122 \(\pm\) 0.0295}	&    \ul{0.0450 \(\pm\) 0.1028}	\\
connect-4        	& \B     0.4181 \(\pm\) 0.0685	&        ---                  	&        0.6831 \(\pm\) 0.1107	&        0.6989 \(\pm\) 0.0745	\\
jungle\_chess\_2pc	& \B     0.1786 \(\pm\) 0.0350	&        0.2390 \(\pm\) 0.0106	&        ---                  	&        ---                  	\\
mfeat-factors    	&        0.1061 \(\pm\) 0.0388	&        0.1160 \(\pm\) 0.0488	& \B     0.0615 \(\pm\) 0.0387	&    \ul{0.1122 \(\pm\) 0.0533}	\\
mfeat-morphologic	&        0.6183 \(\pm\) 0.0687	&        0.6560 \(\pm\) 0.1035	& \B     0.5599 \(\pm\) 0.0780	&        0.5916 \(\pm\) 0.0821	\\
segment          	&    \ul{0.0737 \(\pm\) 0.0692}	&    \ul{0.0596 \(\pm\) 0.0379}	& \B     0.0511 \(\pm\) 0.0293	&    \ul{0.0696 \(\pm\) 0.0362}	\\
vehicle          	&        0.4525 \(\pm\) 0.0495	&    \ul{0.4407 \(\pm\) 0.0898}	& \B     0.3889 \(\pm\) 0.0576	&    \ul{0.4329 \(\pm\) 0.0378}	\\

	\bottomrule
\end{tabular}

\end{table}

It is apparent that, on average, \name{dswizard} outperforms the other frameworks. However, absolute performance differences are small and, especially for \textit{logloss}, often not significant. \name{TPOT} and \name{dswizard}/\name{dswizard*} struggled with some data sets. For \name{TPOT}, single configuration evaluations often exceeded the global timeout leading to an aborted evaluation.  In contrast, \name{dswizard}/\name{dswizard*} exceeded the available memory leading to a crash. The results for \name{dswizard*} show that meta-learning is able to significantly boost the results of \name{dswizard} for \(16\) of \(28\) data sets. Yet, even without meta-learning the more thoroughly evaluation of the pipeline search space yielded configurations outperforming either \name{TPOT} or \name{autosklearn} on \(13\) data sets.

\begin{figure}
    \centering
    \begin{subfigure}[b]{0.49\textwidth}
         \centering
         \includegraphics[width=\textwidth]{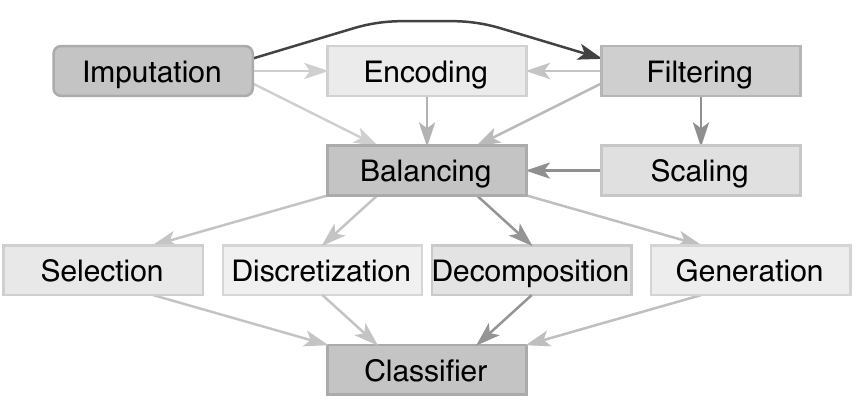}
         \caption{\name{autosklearn}}
         \label{fig:autosklearn_structure}
     \end{subfigure}
     \begin{subfigure}[b]{0.49\textwidth}
         \centering
         \includegraphics[width=\textwidth]{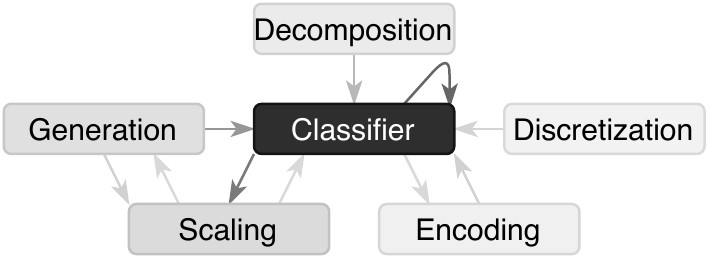}
         \caption{\name{TPOT}}
         \label{fig:tpot_structure}
     \end{subfigure}
     
     \begin{subfigure}[b]{0.49\textwidth}
         \centering
         \includegraphics[width=\textwidth]{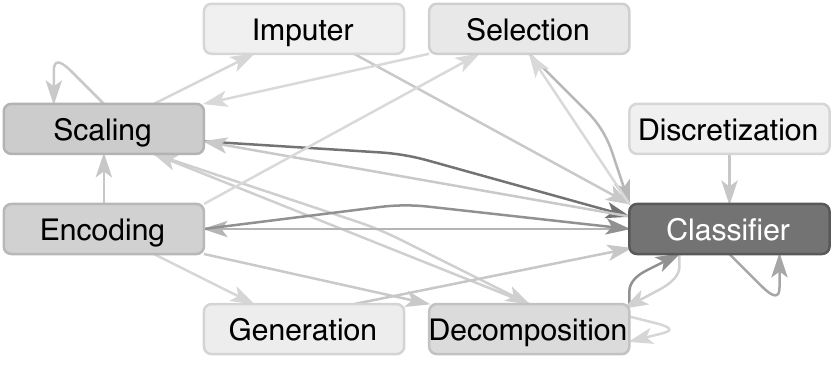}
         \caption{\name{dswizard}}
         \label{fig:dswizard_structure}
     \end{subfigure}
      \begin{subfigure}[b]{0.49\textwidth}
         \centering
         \includegraphics[width=\textwidth]{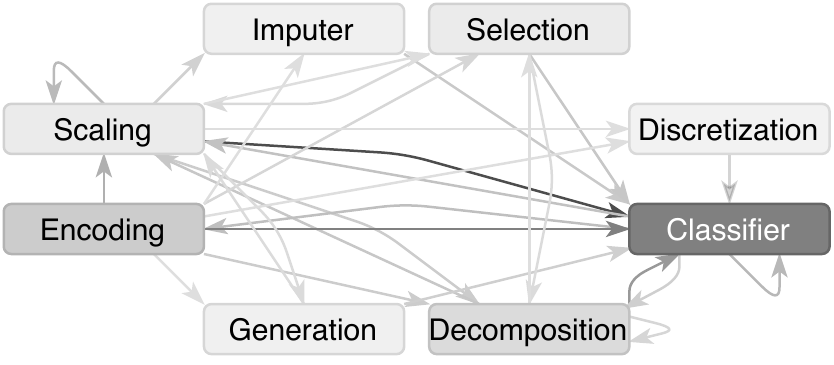}
         \caption{\name{dswizard*}}
         \label{fig:dswizard_star_structure}
     \end{subfigure}

     \caption{Schematic representation of final structures}
     \label{fig:structures}
\end{figure}

Fig.~\ref{fig:structures} shows the structure of the final pipelines aggregated over all data sets and folds. For better readability we substituted each used algorithm by an abstract algorithm class, namely \textit{balancing}, \textit{classifier}, \textit{decomposistion}, \textit{discretization}, \textit{encoding}, \textit{filtering}, \textit{generation}, \textit{imputation}, \textit{scaling} and \textit{selection}. The assignment of algorithm to algorithm class is available in the online Appendix. Additionally, we treat ensembles of pipelines as sets of individual pipelines. Possible pipeline starts are indicated by rounded corners. The frequency of node and edge visits is encoded using a color scheme. Darker colors represent a more frequent usage. For better readability, edges and nodes that appear in less than \(1\%\) of all pipelines are excluded.

In Fig.~\ref{fig:autosklearn_structure} it is clearly visible that \name{autosklearn} uses a semi-fixed pipeline structure where single steps can be omitted. A strict order of algorithms exists, from the top left of the graph to the bottom. \textit{Imputation}, \textit{balancing} and \textit{classifier} are present in each pipeline and the remaining algorithms are roughly selected with identical frequencies. On average, each pipeline contains \(5.48\) algorithms. Due to the semi-fixed structure of \name{autosklearn}, often ineffective algorithms are present in pipelines, \eg imputation even though the data set does not contain missing values. \name{TPOT} is able to construct pipelines with highly varying shapes. However, each pipeline, on average, only contains \(1.66\) algorithms, mostly classifiers. Even though more complex pipelines are constructed, those complex pipelines represent less than \(15\%\) of all pipelines leading to the heavily pruned graph depicted in Fig.~\ref{fig:tpot_structure}. \name{dswizard} constructs more diverse pipelines in terms of selected algorithms and transitions between algorithms. Without meta-learning (compare Fig.~\ref{fig:dswizard_structure} and \ref{fig:dswizard_star_structure}), structure search is less guided leading to longer pipelines (\(2.85\) vs. \(3.07\) algorithms) and a more uniform distribution of the selected algorithms. Yet, as \name{dswizard*} performs worse than \name{dswizard}, the guidance seems to be helpful to find well performing pipeline structures.

Finally, we take a more detailed look at \textit{credit-g} and \textit{higgs}, two data sets where \name{dswizard} performs especially well and badly, respectively. For \textit{credit-g} \(71\)\% of the pipelines in the final ensemble have a long complex structure that can not be created by \name{autosklearn}. Most pipelines combine stacked classifiers with multiple preprocessors. For \textit{higgs} the meta-learning directs the structure search in a wrong direction leading to many ineffective transformations. As a result, only very basic pipelines containing combinations of the same four algorithms are constructed. Even with \ac{HPO}, these simple pipelines do not perform well.

In summary, \name{dswizard} significantly outperforms either \name{autosklearn} or \name{TPOT} on \(~42\)\% of the data sets. Moreover, it has a similar performance to the state-of-the-art on \(~32\)\% of the data sets.

\section{Conclusion}
\label{sec:conclusion}
We presented a data-centric approach for solving the \ac{PSO} problem inspired by human behaviour using \ac{MCTS} in combination with Bayesian optimization. The possibility to expand the search space incrementally allows for a dynamic adaptation of the pipeline structure to a specific data set. Unpromising regions of the pipeline search graph can be identified and discarded quickly---often even without \ac{HPO}---through the extensive use of meta-features allowing an efficient traversal of the growing search space. Furthermore, sharing knowledge between pipeline structures is implemented for warm-starting \ac{HPO}. This allows a more thorough exploration of the pipeline structure search space with a dynamic adaptation to a specific data set while still obtaining competitive results.

\bibliographystyle{splncs04}
\bibliography{library}

\end{document}